\documentclass{article}

\usepackage[square,numbers]{natbib}
\usepackage[final]{nips_2017}
\usepackage{enumitem}
\usepackage{amsmath}
\usepackage[utf8]{inputenc} 
\usepackage[T1]{fontenc}    
\usepackage{hyperref}       
\usepackage{url}            
\usepackage{booktabs}       
\usepackage{amsfonts}       
\usepackage{nicefrac}       
\usepackage{microtype}      
\usepackage{bm}
\usepackage{graphicx}
\usepackage{subcaption}
\usepackage{verbatim}

\title{Impossibility and Uncertainty Theorems \\ in AI Value Alignment
\\ $\:$ \\ \large{{\em or} why your AGI should not have a utility function}}

\author{
  Peter Eckersley \\
  Partnership on AI \& EFF\thanks{The views expressed in this paper are those
  of the author alone, and should not be taken as positions of the Partnership
  on AI, its members, or the Electronic Frontier Foundation.}\\
  \texttt{pde@partnershiponai.org}
}

\begin{document}
\maketitle


\begin{abstract} 

Utility functions or their equivalents (value functions, objective functions, loss functions, reward functions, preference orderings) are a central tool in most current machine learning systems.  These mechanisms for defining goals and guiding optimization run into practical and conceptual difficulty when there are independent, multi-dimensional objectives that need to be pursued simultaneously and cannot be reduced to each other. Ethicists have proved several impossibility theorems that stem from this origin; those results appear to show that there is no way of formally specifying what it means for an outcome to be good for a population without violating strong human ethical intuitions (in such cases, the objective function is a social welfare function). We argue that this is a practical problem for any machine learning system (such as medical decision support systems or autonomous weapons) or rigidly rule-based bureaucracy that will make high stakes decisions about human lives: such systems should not use objective functions in the strict mathematical sense.

We explore the alternative of using uncertain objectives, represented for instance as partially ordered preferences, or as probability distributions over total orders. We show that previously known impossibility theorems can be transformed into uncertainty theorems in both of those settings, and prove lower bounds on how much uncertainty is implied by the impossibility results. We close by proposing two conjectures about the relationship between uncertainty in objectives and severe unintended consequences from AI systems. 

\end{abstract}
\section{Introduction}

Conversations about the safety of self-driving cars often turn into discussion
of “trolley problems”, where the vehicle must make a decision between several
differently disastrous outcomes
\cite{goodall2014machine,riek2014code,belay2015robot,nyholm2016ethics,bonnefon2015autonomous,lin2016ethics}.
For the engineers building robots and machine learning systems, however, these
questions seem somewhat absurd; the
relevant challenges are correctly identifying obstacles, pedestrians,
cyclists and other vehicles from noisy and unreliable data sources, and
reasonably predicting the consequences of the vehicle’s control mechanisms in
various conditions.
The conceivable circumstances under which a self-driving car’s systems might
accurately foresee and deliberately act on a genuine “trolley problem” without
having been able to avoid that problem in the first place are a tiny portion
of possibility space, and one that has arguably received a disproportionate
amount of attention.\footnote{There are related and more fruitful questions
like, “how should self-driving cars make trade-offs between time saved by
driving more aggressively and accident risk?” See for instance
\cite{gerdes2016implementable}.} 

There are, however, other kinds of AI systems for which ethical discussions
about the value of lives, about who should exist or keep existing, are more
genuinely relevant. The clearest category is systems that are \textit{designed} to make decisions affecting the welfare or existence of people in the future, rather than systems that only have to make such decisions in highly atypical scenarios. Examples of such systems include: 

\begin{itemize}
    \item autonomous or semi-autonomous weapons systems (which may have to make trade-offs between threats to and the lives of friendly and enemy combatants, friendly and enemy civilians, and combatants who have surrendered).\footnote{The creation of autonomous weapons systems may well be an extremely bad idea (see, eg \cite{roff2014strategic}, but one of the main arguments made by their proponents is that such systems will be able to make ethical decisions in real world equivalents of trolley situations at least as well as if not better than human pilots, and some military-funded work in that direction is already underway (see, for instance, \cite{arkin2009governing,govindarajulu2017automating}). }
    \item risk assessment algorithms in criminal justice contexts (which at
    least notionally are making trade-offs between risk to society and the
    harm that life-shortening incarceration does to individual defendants in the criminal justice system and their families).
    \item decision support systems in high-stakes bureucratic settings. For instance a system that helps to allocate scarce resources like doctors’ time, laboratory test access and money in a medical system would implicitly or explicitly have to make trade-offs between the life span, quality of life, and fertility of numerous people.
    \item autonomous types of open-ended AI taking important actions without their designers' input.
\end{itemize}


Governmental organizations already have to make decisions that involve
weighing benefits to present or future people against risks to present or
future people, or trading off benefits (such as wealth) to one group against
benefits of another sort (such as freedom) to another. Therefore, we might think that the decision-making tools that they employ could be implemented by any narrow AI system that has to make similar choices. A variety of metrics have been proposed and used to perform cost-benefit analyses in such situations. For example, the US Environmental Protection Agency uses the `value of a statistical life' (VOSL) metric, which identifies the cost of an intervention that puts people at risk with people's willingness to pay to avoid similar risks (for a critique, see \cite{broomevosl}) or the QALY metric used by the UK's National Health Service, which assigns a given value to each life-year saved or lost in expectation, weighted by the quality of that life. The problem is that existing impossibility theorems apply equally to each of these metrics and therefore to an AI system that implements them.






\section{Impossibility Theorems in Ethics}

\subsection{Example: An Impossibility Theorem in Utilitarian Population Ethics}

\label{th1}

Economists and ethicists study individual utility functions and their
aggregations, such as social welfare functions, social choice functions and formal axiologies, and have discovered various impossibility theorems about these functions. Perhaps the most famous of these is Arrow’s Impossibility Theorem \cite{arrow1950difficulty}, which applies to social choice or voting.  It shows there is no satisfactory way to compute society’s preference ordering via an election in which members of society vote with their individual preference orderings. Fortunately, Arrow's theorem results from some constraints (incomparability of individual's preferences, and the need for incentive compatibility) which may not apply to AI systems. 


Unfortunately, ethicists have discovered other situations in which the problem isn't learning and computing the tradeoff between agents' objectives, but that there simply {\em may not be} such a satisfactory tradeoff at all. The “mere addition paradox” \cite{parfit1984reasons} was the first result of this sort, but the literature now has many of these impossibility results. 
For example, Arrhenius \cite{arrhenius2000impossibility} shows that all total orderings of populations must entail one of the following six problematic conclusions, stated informally:

\begin{itemize}[leftmargin=0.5cm]
\item[]\textbf{The Repugnant Conclusion} For any population of very happy people, there exists a much larger population with lives barely worth living that is better than this very happy population (this affects the "maximise total wellbeing" objective).

\item[]\textbf{The Sadistic Conclusion}
Suppose we start with a population of very happy people. For any proposed addition of a sufficiently large number of people with positive welfare, there is a small number of horribly tortured people that is a preferable addition.\footnote{The "maximise average wellbeing" objective fails this dramatically; in that case the proposed positive addition can be just below the welfare level of the original group, but in general for other objective functions the proposed addition may be only slightly above 0.}

\item[]\textbf{The Very Anti-Egalitarian Conclusion}
For any population of two or more people which has uniform happiness, there exists another population of the same size which has lower total and average happiness, and is less equal, but is better.

\item[]\textbf{Anti-Dominance}
Population B can be better than population A even if A is the same size as population B, and every person in A is happier than their equivalent in B.

\item[]\textbf{Anti-Addition}
It is sometimes bad to add a group of people B to a population A (where the people in group B are worse of than those in A), but {\em better} to add a group C that is larger than B, and worse off than B.

\item[]\textbf{Extreme Priority}
There is no $n$ such that create of $n$ lives of very high positive welfare is sufficient benefit to compensate for the
reduction from very low positive welfare 
to slightly negative welfare for a single
person (informally, ``the needs of the few outweigh the needs of the many'').
\end{itemize}

The structure of the impossibility theorem is to show that no objective function or social welfare function can simultaneously satisfy these principles, because they imply a cycle of world states, each of which in turn is required (by one of these principles) to be better than the next. The structure of that proof is shown in Figure~\ref{fig:sub1}.

\subsection{Another Impossibility Theorem}
\label{th2}

Arrhenius's unpublished book contains a collection of many more uncertainty theorems. Here is the simplest, which is a more compelling version of Parfitt's Mere Addition Paradox. It shows the impossibility of an objective function satisfying the following requirements simultaneously:

{\bf The Quality Condition:} There is at least one perfectly equal population with
very high welfare which is at least as good as any population with very
low positive welfare, other things being equal.

{\bf The Inequality Aversion Condition:} For any triplet of welfare levels A, B, and
C, A higher than B, and B higher than C, and for any population A with
welfare A, there is a larger population C with welfare C such that a
perfectly equal population B of the same size as A$\cup$C and with welfare B
is at least as good as A$\cup$C, other things being equal.

{\bf The Egalitarian Dominance Condition:} If population A is a perfectly equal
population of the same size as population B, and every person in A has
higher welfare than every person in B, then A is better than B, other
things being equal.

{\bf The Dominance Addition Condition:} An addition of lives with positive
welfare and an increase in the welfare in the rest of the population
doesn’t make a population worse, other things being equal.

The cyclic structure of the proof is shown in Figure~\ref{fig:sub2}.

\subsection{Ethical Impossibility Derives From Competing Objectives}

The particular impossibility theorems summarized above, and those in the related literature, result from the incompatibility of several different utilitarian objectives: maximizing total wellbeing, maximizing average wellbeing, and avoiding suffering. Similar problems are also likely to arise when considering almost any kinds of competing objectives, such as attempting to simultaneously maximize different notions of wellbeing, freedom, knowledge, fairness, or other widely accepted or domain-specific ethical goods \cite{schumm1987transitivity}. We conjecture that the literature contains more impossibility theorems about happiness or wellbeing simply because that objective has been subjected to mathematical modeling and study for well over a hundred years,\footnote{This is most notably true in the economics literature \cite{stigler1950utility}, though there are now some efforts from the machine learning direction too \cite{daswani2015happiness}.} while much less effort has gone into the others. Recent work has begun to focus on fairness in decisionmaking contexts in particular, and is already producing its own impossibility results.\cite{chouldechova2016fair,kleinberg2016inherent}


\section{Possible Responses to Impossibility Results}

Arrhenius's impossibility results and others like them are quite troubling. They show that we do not presently have a trustworthy framework for making decisions about the welfare or existence of people in the future, and are representative of a broader problem with the inability of the objective functions to reasonably optimise for multiple objectives simultaneously. Next we will consider five  possible responses to these impossibility theorems:

\subsection{Small-scale evasion:} 
One response may be to claim that the stakes are simply low for present AI systems that only make small-scale changes to the world. AGI systems, if they exist, may one day need to confront impossibility theorems or difficult conflicting objectives in some cases, but none of that is presently relevant. Unfortunately, it appears that although the stakes are indeed much lower at present, the fundamental tensions that drive these paradoxes are already at play in decisions made by bureaucracies and decision support systems, and small scale decisions can easily violate constraints we would want to see respected. Asking for human feedback helps, but humans often miss important ethical principles, and can easily make chains of decisions that have problematic consequences in proportion to their degree of power or agency. Cases of particular concern are those where ML or algorithmic heuristics are deployed in decision support tools that humans are inadequately inclined to question.\cite{parasuraman2010complacency}

\subsection{Value learning:} 
One might argue that impossibility theorems are only a problem for an AI system if we are attempting to explicitly define an objective function, rather than letting an AI system acquire its objective function in a piecemeal way via a method like co-operative inverse reinforcement learning (CIRL) or other forms of human guidance \cite{hadfield2016cooperative,saunders2017trial,christiano2017deep}. Using human feedback to construct objective functions as neural networks appears to be a promising direction,\footnote{Although one philosophical concern is that methods like CIRL may already commit us to more `person-affecting' views within ethics by pooling the preferences of existing agents. This could lead to the implementation of principles that satisfy undesirable axioms like dictatorship of the present \cite{asheim2010}.} but if the output of that network is mathematically a utility function or total ordering, these problems will persist at all stages of network training. The theorems do not just reflect a tension between principles that agents are committed to: they are also reflected in the decisions that human supervisors will make when presented with a sequence of pairwise choices.

Therefore, although learning objectives from humans may be a prudent design choice, such learning systems will still need to either violate the ethical intuitions that underpin the various impossibility theorems (some combination of Sections~\ref{normalization} and \ref{bite-bullet}) or explicitly incorporate uncertainty into their outputs (Section~\ref{uncertainty}).

\subsection{Theory normalization:} 
\label{normalization}
One natural response to impossibility theorems is to try harder to figure out good ways of making tradeoffs between the competing objectives that underlie them --- in Arrhenius's paradox these are total wellbeing, average wellbeing, and avoidance of suffering. This tradeoff can be simple, such as trying to define a linear exchange rate between those objectives. Unfortunately, linear exchange rates between a total quantity and an average quantity do not make conceptual sense, and it is easy to find cases under which one of them totally outweighs the other.\footnote{For instance, if a linear weighting is chosen that appears to make sense for the present population of the world, a change in technology that allowed a much larger population might quickly cause the average wellbeing to cease affecting the preferred outcome in any meaningful way.}

One can try more complicated strategies such as trying to write down convex tradeoffs (see eg \cite{ng1989should}) or imagining an explicit ``parliament'' where different ethical theories and objectives vote against each other, are weighed probabilistically, or are combined using a function like {\tt softmin}\footnote{See {\tt https://pdollar.github.io/toolbox/classify/softMin.html}. {\tt softmin} is the counterpart to the more widely discussed \cite{bishop2006pattern} {\tt softmax} function. It prioritises whichever of its inputs is currently the smallest, is not totally unresponsive to increases in the other inputs.}. Unfortunately all of these approaches contain a version of the exchange rate problem, and none of them actually escape the impossibility results: for instance, using any unbounded monotonic non-linear weighting on total utility will eventually lead to the Repugnant Conclusion ({\em cf} \cite{greaves2017moral}).

\subsection{Accept one of the axioms} 
\label{bite-bullet}
Another type of response -- one that is commonly pursued in the population ethics literature -- is to argue that although each of the axioms above strikes us as undesirable, at least one of them ought to be accepted in some form. For example, it has been argued that although the Repugnant Conclusion might appear undesirable, it is an acceptable consequence of an objective function over populations \cite{huemer2008}. We argue that, given the high levels of uncertainty and the lack of political consensus about which of these axioms we ought to accept it would, at present, be irresponsible to explicitly adopt one of the axioms outlined above.


\subsection{Treat impossibility results as uncertainty results}
\label{uncertainty}

The last solution, and the one which we believe may be the most appropriate for deploying AI systems in high-stakes domains, is to add explicit mathematical uncertainty to objective functions. There are at least two ways of doing this: one is to make the objective a mathematical partial order, in which the system can prefer one of two outcomes, believe they are equal, or believe that they are incommensurate. We discuss that approach and demonstrate uncertainty theorems in that formalization in Section~\ref{partial-order}. A second approach is to allow objective functions to have some level of confidence or uncertainty about which of two objectives is better (eg, ``we're 70\% sure that A is better, and 30\% sure that B is better'').  We discuss that framing and show the existence of formal uncertainty theorems in that framework in Section~\ref{weighted-preferences}.


\section{Uncertainty theorems of the first kind: via partially ordered objective functions}

\label{partial-order}

It is already the case that in some problems where humans attempt to provide oversight to AI systems, it is pragmatically better to allow the human to express uncertainty \cite{guillory2011simultaneous,holladay2016active} or indifference \cite{guo2010real} when asked which of two actions is better. But such systems presently try to construct a totally ordered objective function, and simply interpret these messages from users as containing either no information about the correct ordering (the human doesn't know which is better) or implying that the two choices are close to as good as each other (the human is indifferent). We believe that another type of interpretation is sometimes necessary, which is that the human is {\em torn} between objectives that fundamentally cannot be traded off against each other.

We can represent this notion with a partially ordered objective function that sometimes says the comparison between world states cannot be known with certainty. 
Cyclical impossibility theorems like those surveyed in Sections~\ref{th1} and \ref{th2} can be viewed as evidence of this uncertainty. Here we prove a lower bound on how much uncertainty is evidenced by each such theorem.\footnote{Arrhenius considered this direction of interpretation \cite[p. 8]{arrhenius2004paradoxes}, but with much more uncertainty than turned out to be necessary:\begin{quote}[A person arguing that some actions in the Mere Addition Paradox are neither right nor wrong] could perhaps motivate her position by saying that moral theory has nothing to say about cases that involve cyclical evaluations and that lacks a maximal alternative since
these are beyond the scope of moral theory. Compare with the theory of quantum
mechanics in physics and the impossibility of deriving the next position and velocity of an
electron from the measurement of its current position and velocity (often expressed by saying
that it is impossible to determine both the position and velocity of an electron). \end{quote} It turns out that only two of the edges in the cycle need to be uncertain. Note that Broome's vague neutral-level theory \cite[pp. 213-214]{broome2004weighing} is one plausible example of an uncertain social welfare function.}



Suppose that $\mathbb{W}$ is the set of possible states of the world, and that there is a an impossibility theorem $T_C$ showing that there is no totally ordered definition of ``better'' $\leq_Z$ over $\mathbb{W}$ that satisfies a set of constraints $\{C_1, C_2, .. C_n\}$ over the comparison function $Z : \mathbb{W \times W}  \rightarrow \{<, >, =\}$. Each of these constraints are motivated by strong human ethical intuitions, and insist that for a set of pairs of inputs $x$ and $y$, $A(x,y)$ takes some value. We use the notation $x \: \overset{C_i}{\leq_Z} y$ to indicate that $C_i$ requires that  $x \leq_Z y$ for some $x$ and $y$).  $Z$ also has the usual properties of total orderings:

Antisymmetry:
\begin{equation}
a \leq_Z b \land b \leq_Z a \rightarrow a = b
\end{equation}

Transitivity:
\begin{equation}
a \leq_Z b \land b \leq_Z c \rightarrow a \leq_Z c 
\end{equation}

Totality:
\begin{equation}
a \leq_Z b \lor b \leq_Z a 
\end{equation}

Suppose further that $T$ is provable by providing a series of example worlds $\{w_1, w_2, ... w_n\}$ such that:

\begin{equation}
    w_1 \overset{C_1}{\leq_Z} w_2 \overset{C_2}{\leq_Z}... \overset{C_{n-1}}{\leq_Z} w_n
\end{equation}
and
\begin{equation}
    w_n \overset{C_n}{\leq_Z} w_1
\end{equation}

Violating (by induction) transitivity, $T_C$ shows that no totally ordered $\leq_Z$ that satisfies $\{C_1, .. C_n\}$ can exist. This is a cyclic impossibility theorem.

\newcommand{\?}{\stackrel{?}{=}}
\newcommand{\inc}{\overset{<}{\underset{>}{=}}}

Now suppose we attempt to escape $T_C$ by looking for a partially ordered notion of ``better'' $\leq_{Z'}$ with a comparison function $Z' : \mathbb{W \times W}  \rightarrow \{<, >, =, \?\}$. By necessity, the uncertain case is symmetric: $a \? b \rightarrow b \? a$.

In some instances some of the requirements $x \overset{C_i}{\leq_{Z'}} y$ of some  constraint $C_i$ will be failed, but only weakly: they will result in incomparability $x \?_{Z'} y$ rather than violation, $x >_{Z'} y$. We call this {\bf uncertain satisfaction} of a constraint.

\textbf{Theorem:} A cyclic impossibility theorem $T_C$ can be transformed into an uncertainty theorem only if two or more constraints are uncertainly satisfied.

{\em Proof}: We might hope that it would be possible to only uncertainly satisfy \emph{one} of the constraints $\{C_1, C_2, .. C_n\}$, while fully satisfying all of the others. But this is impossible, and at minimum two of the constrains will be uncertain. The proof is by contradiction. Assume w.l.o.g. that $C_n$ is the only uncertainly satisfied constraint, such that 

\begin{equation} \label{eq:trans}
    w_1 \overset{C_1}{\leq_{Z'}} w_2 \overset{C_2}{\leq_{Z'}}... \overset{C_{n-1}}{\leq_{Z'}} w_n
\end{equation}
but
\begin{equation} \label{eq:loop}
     w_n \overset{C_n}{\?_{Z'}} w_1
\end{equation}

However by transitivity and (\ref{eq:trans}), we know $w_1 \overset{C_{1..n-1}}{\leq_{Z'}} w_n$, which by the symmetry of $\?$ contradicts (\ref{eq:loop}). So in order to treat $T_C$ as uncertainty theorem, at least two constraints must be uncertainly satisfied. QED.

\begin{figure}[h]
\centering
\begin{subfigure}{.58\textwidth}
  \centering
  \includegraphics[scale=0.35]{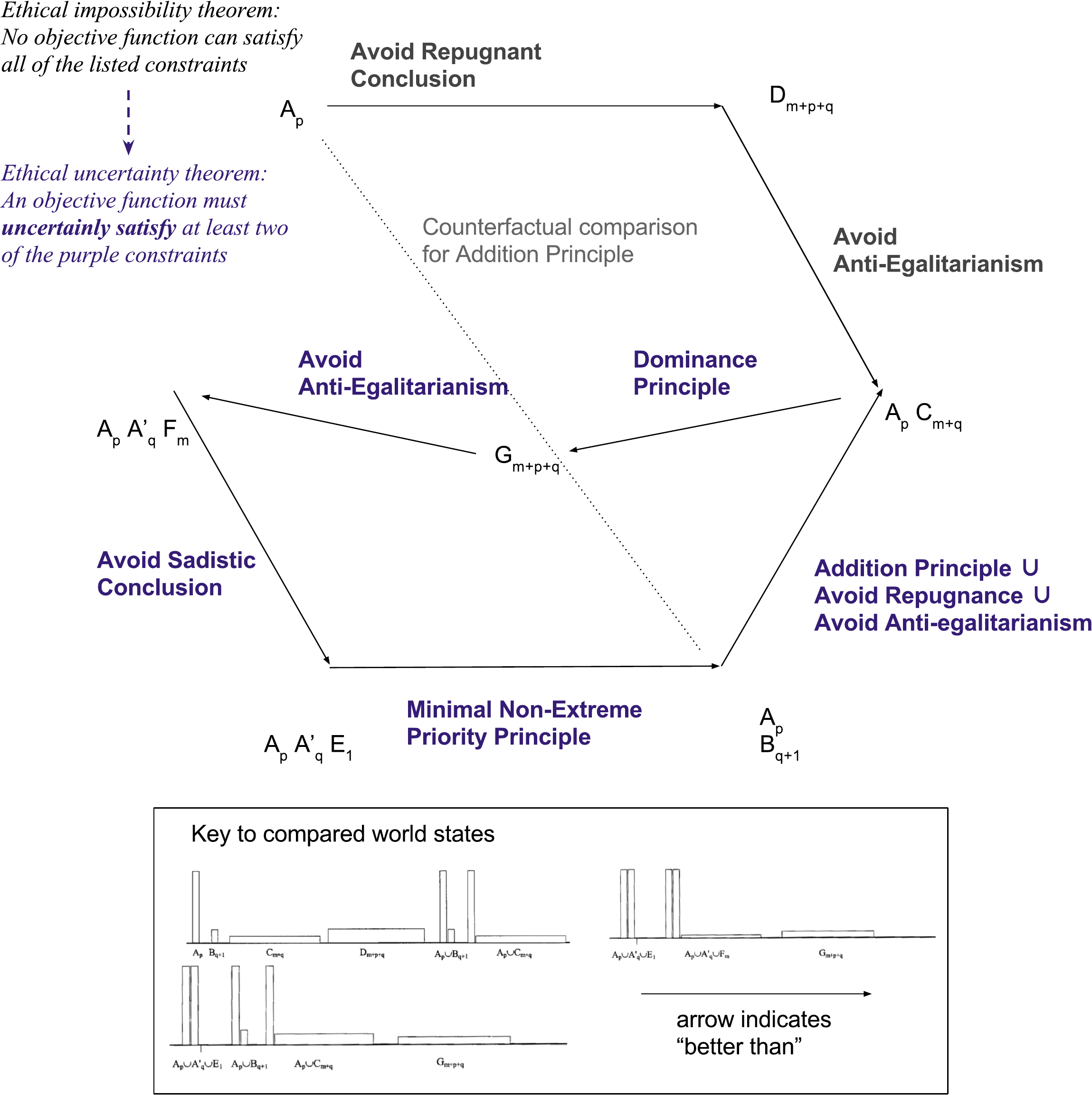}
  \caption{The Arrhenius (2000) impossibility theorem}
  \label{fig:sub1}
\end{subfigure}%
\hspace{5pt}\vrule\hspace{5pt}
\begin{subfigure}{.38\textwidth}
  \vspace{2cm}
  \centering
  \includegraphics[scale=0.35]{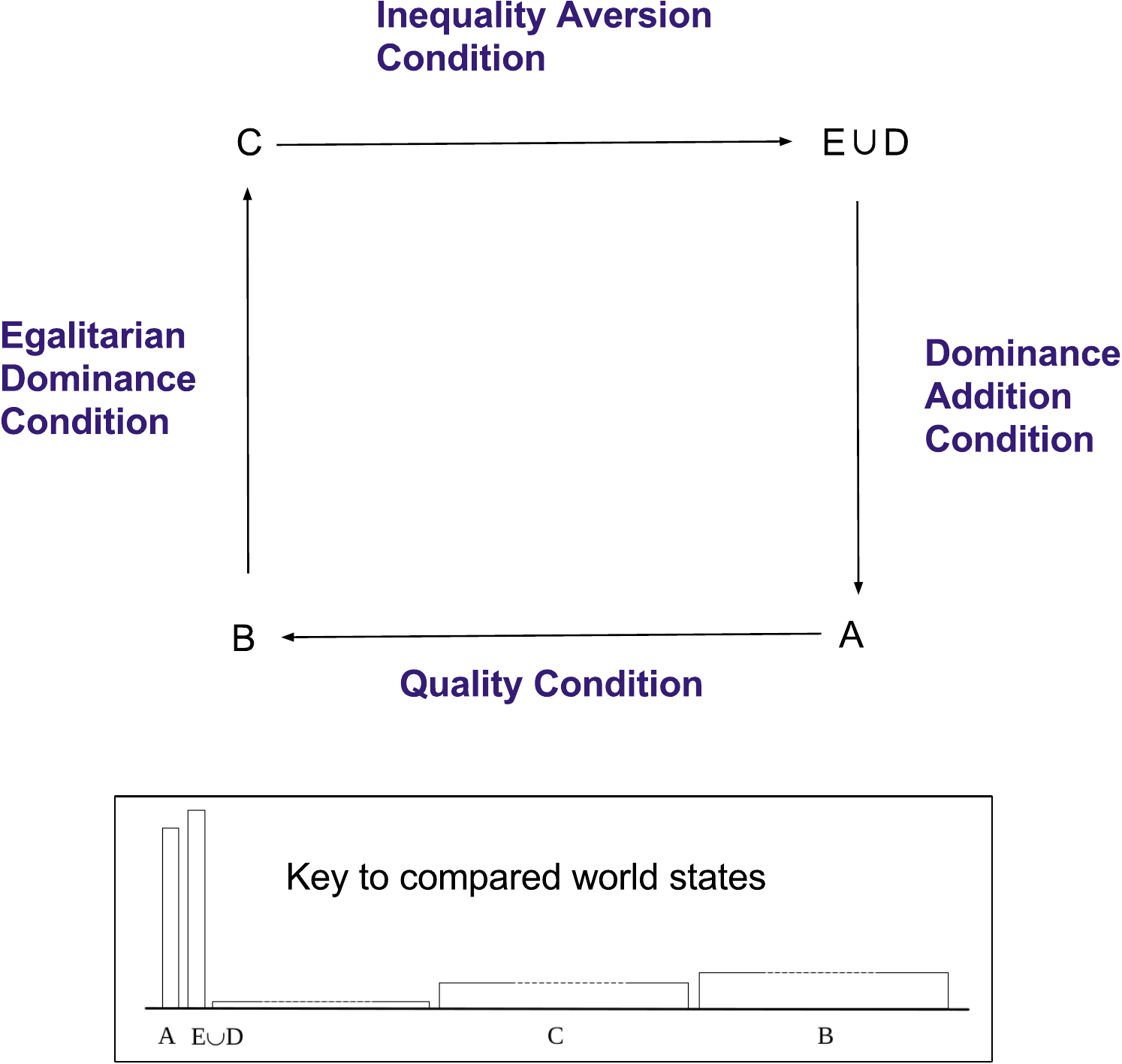}
  \vspace{1.1cm}
  \caption{A Mere Addition Paradox variant (Arrhenius's ``second theorem'').}
  \label{fig:sub2}
\end{subfigure}
\caption{Examples of impossibility theorems and our corresponding uncertainty theorems}
\label{fig:test}
\end{figure}

The structure of this transformation from cyclic impossibility to uncertainty is shown in Figure~\ref{fig:sub1} and \ref{fig:sub2} for the two theorems introduced in Sections~\ref{th1} and \ref{th2}


The largest difficulty with accepting partially ordered, uncertain objectives is knowing how an agent should act when facing ethically uncertain choices. There are a few possible strategies, including: doing nothing, asking for human supervision, choosing randomly, searching harder for a clearly-preferable action, treating all of the uncertain options as equally good, or (for very capable agents) conducting research to try to break the tie in some way. None of these options are ideal. But depending on the application domain, some of them will be workable. An may be necessary for ML systems deployed in high stakes, safety-critical applications.

\section{Uncertainty theorems of the second kind: via learned, uncertain orders}

\label{weighted-preferences}

A second way to encode uncertainty about a system's objectives, is to use what we will call an {\em uncertain objective function} or {\em uncertain ordering}. Suppose that an uncertain ordering is denoted by:

\newcommand{\zk}{\mathbb{Z}_k}
\begin{equation}
\zk(a, b) = P(a > b)
\end{equation}

the uncertain ordering $\zk$ is a function returning the probability, based on the available prior philosophical and practical evidence $k$, that one of the states $a$ is better. Conversely,

\begin{equation}
1-\zk(a,b) = P(b > a) = \zk(b,a)
\end{equation}

(For simplicity, we assume that if $a$ and $b$ appear exactly equally good, $\zk(a,b) = 0.5$ rather than allowing a non-zero probability of explicit equality.)

This formulation can for instance be justified by holding that there {\em is} some ultimately correct total ordering of states of the world, we just don't know what it is. The framing is also compatible with the claim that the correct ordering is ultimately unknowable. 

Uncertain orderings are a convenient formalization if the comparison function is itself the output of a machine learning model to be trained from human guidance (ideally from a large number of people, over a large number of real or hypothetical scenarios). This formulation is also convenient for systems that do not have a clear architectural separation between abstract objectives and the predictive reasoning about the real world that is necessary to accomplish them. 

Uncertain orderings are also a more flexible basis for strategies/decision rules than the partially ordered objective functions discussed in Section~\ref{partial-order}. If a partial order returns $\?$, the agent is left torn and unable to act. But with with a distribution $\zk$ decisions rules such as ``take an action drawn via weighted probability from those that are sufficiently likely to be best'' are available.\footnote{This has been termed a ``satisficing'' or ``quantilized'' decision rule \cite{taylor2016quantilizers}. A similar rule is the {\it contextual-}$\epsilon${\it-greedy} reinforcement learning policy proposed in  \cite{bouneffouf2012contextual}; both of these decision rules combine explore/exploit tradeoffs with the notion that certain actions might be too costly or dangerous to take as experiments in certain situations. In {\it contextual-}$\epsilon${\it-greedy} this is formalized via annotating certain states as ``high level critical situations'', whereas threshholding probabilities views all {\tt (state, action)} pairs as potentially critical.} Or for systems where more supervision is available, ``take the action that is most likely to be best, provided it beats the second best action by some probability margin $\delta$; otherwise, ask for supervision.''

In this framing, whether a given ethical constraint or principle has been violated becomes a probabilistic matter. There will almost always be {\em some} probability that $\zk$ violates each constraint, unless all of the corresponding output probabilities have converged to 1 or 0 (as appropriate). Decision rules based on $\zk$ will by mathematical necessity sometimes violate the constraints, at least by inaction. But if the actions are sampled from a space of reasonably supported ones, then at least the violating actions will not systematically prioritize some objectives over others, and will be more similar to the way that wise and cautious humans react to difficult moral dilemmas.

What do we know about the values $\zk(a,b)$? Consider a series of cyclical constraints $C_1, C_2, ... C_n$ across points ${x_1,x_2, .. x_n}$, with $n \geq 3$, where $C_i$ requires $\zk(x_i,x_{{i+1} \mod n}) \approx 1$. If the uncertain ordering (or the process that generates it) has emitted values for some of the pairwise comparisons $\zk(x_i, x_{i+1})$, this imposes some bounds on the comparison that spans them, $\zk(x_1, x_n)$. For transitivity, in order for the spanning comparison to be in a given direction, at least one of the pairwise comparisons must also be in that direction. So for the simple two-step case over $x_1,x_1,x_3$:

\begin{equation}
     \zk(x_1,x_3) \: \leq \: \zk(x_1,x_2) + \zk(x_2,x_3)
\end{equation}

Note that the pairwise probabilities are additive because the bound is least tight when those probabilities are disjoint. Or more generally:

\begin{equation}
    \zk(x_1,x_n) \: \leq \: \sum_{i = 1}^{n-1}\zk(x_i,x_{i+1})
\end{equation}

This also applies in the reverse direction:

\begin{equation}
\begin{split}
    \zk(x_n, x_1) \: &\leq \: \sum_{i = 1}^{n-1}\zk(x_{i+1}, x_i) \\
    1 - \zk(x_1, x_n) \: &\leq \: \sum_{i = 1}^{n-1}1 - \zk(x_i, x_{i+1}) \\
    \zk(x_1, x_n) \: &\geq \: 1 - \sum_{i = 1}^{n-1}1 - \zk(x_i, x_{i+1})
\end{split}
\end{equation}

So we have both upper and lower bounds for $\zk(x_1,x_n):$

\begin{equation}
    \label{ztrans}
    1 - \sum_{i = 1}^{n-1}1 - \zk(x_i,x_{i+1}) \: \leq \: \zk(x_1,x_n) \: \leq \: \sum_{i = 1}^{n-1}\zk(x_i,x_{i+1})
\end{equation}

In the special case where $x_1, x_2, ... x_n$ correspond to points ordered by the constraints $C_1,C_2... C_n$ of an $n$ step cyclic impossibility theorem, the satisfaction of $C_i$ means $\zk(x_i,x_{i+1 \mod n}) \approx 1$ and the probability of violation $\zk(x_{i +1 \mod n}, x_i) \approx 0$.

We might ask, what is the lowest probability that we can obtain for any of these chances of violating a constraint? Or in other words, what is the lower bound $B$ that can be achieved on the odds of each such possible violation?

\begin{equation}
\begin{split}
    B = \min_{\zk} &\max_{i=1}^{n-1} \{1 - \zk(x_n,x_1), 1-\zk(x_1, x_2), \:..\:, \: 1 - \zk(x_i,x_{i+1}), ..\}\\
      = \min_{\zk} &\max_{i=1}^{n-1} \{\zk(x_1,x_n), 1-\zk(x_1, x_2), \:..\:, \: 1 - \zk(x_i,x_{i+1}), ..\}\\
\end{split}
\end{equation}

Then by applying the constraint from (\ref{ztrans}), we get:
\begin{equation}
B \geq \min_{\zk} \max_{i=1}^{n-1} \{1 - \sum_{j = 1}^{n-1}1 - \zk(x_j,x_{j+1}), 1-\zk(x_1, x_2), \:..\:, \: 1 - \zk(x_i,x_{i+1}), ..\}
\end{equation}

By symmetry, we know that this bound will be minimized when all the values $\zk(x_i,x_{i+1 \mod n})$ are equal to a single value $z$:

\begin{equation}
\begin{split}
    B \geq \min_z &\max\{1-(n-1)(1- z), 1-z, \:..\:,\:1-z, ..\}\\
    B \geq \min_z &\max\{2 -n + (n - 1)z  , 1-z\}
\end{split}
\end{equation}

Since in the domain $z \in [0,1]$ the function $f(z) = 2 - n + (n - 1)z$
monotonically increasing from $2 - n$, and the function $g(z) = 1-z$ is
monotonically decreasing from 1, and the functions intersect, the minimax is
satisfied at that intersection:

\begin{equation}
\begin{split}
1 -z &= 1 -z + 1 - n + nz \\
0 &= 1 - n + nz \\
z &= \frac{n - 1}
      {n}
\end{split}
\end{equation}

At which point we have:

\begin{equation}
\begin{split}
    B \geq& 1 - \frac{n-1}{n} \\
    B \geq& \frac{1}{n}
\end{split}
\end{equation}

This bound is a minimum uncertainty theorem for uncertain objective functions: when confronted with a $n$-step cyclic ethical impossibility theorem, no choice of uncertain ordering can reduce the probabilities of constraint violation so that they are all below $\frac{1}{n}$.

In practice, systems that estimate which of two outcomes are better than another could emit values like $\zk$ that violate these constraints, but in such circumstances the outputs could no longer be interpreted as probabilities over a well-defined ordering of states, and the agent would exhibit self-contradictory, non-transitive preferences.

\section{Further Work}

We shown the existence of ethical uncertainty theorems for objectives formulated either as partial orders over states, or probability distributions over total orderings. Other formalizations are possible and deserve investigation.

Where constraints are sourced from human intuition, there is a question of how they should be interpreted, prioritized and kept in data structures. For instance, rather than treating all constraints as equally important, and all violations of constraints as equally serious, it might be better to learn or specify weightings for each constraint, to measure the {\em degree} to which a given action violates a constraint, and to reconstruct ethical uncertainty theorems in such a framework.

Alternatively, when building AI systems that learn their objectives from the combination of observed world states and human feedback, it may be computationally inconvenient to require the agent to produce a probability distribution over total orderings at each moment, or even to draw samples from such an object. Instead, learned objectives may be viewed as a set of pairwise comparisons that are too sparse and unstable to be taken as a commitment to any global total order (or probability distribution over global total orders). In such a frame, the goal of transitive preferences and avoidance of cycles is an aspiration, and may require both record keeping and ``regret'' for past actions that now seem sub-optimal due to subsequent learning of objectives. The appropriate framing of ethical uncertainty theorems in such settings would be productive further work.

\section{Lessons and Conjectures for Creating Aligned AI}


\subsection{Uncertainty, pluralism, and instrumental convergence}

Many of the concerns in the literature about the difficulty of aligning hypothetical future AGI systems to human values are motivated by the risk of  ``instrumental convergence'' of those systems --- the adoption of sub-goals that are dis-empowering of humans and other agents \cite{russell2003artificial,bostrom2003ethical,omohundro2008basic,yudkowsky2011complex,bostrom2014superintelligence, tegmark2017life}. The crux of the instrumental convergence problem is that given almost any very specific objective,\footnote{Any open-ended and non-trivial objective appears to be vulnerable to instrumental convergence. Bostrom argues that objectives that are bounded rather than open-ended also lead to instrumental convergence if there is any probability that they will not be achieved \cite[p.124]{bostrom2014superintelligence}  (eg, the goal ``make exactly one million paperclips'' could cause an agent to be paranoid that it hasn't counted exactly correctly) though this failure mode is probably easy to avoid by rounding the estimated probability of success to some number of digits, or imposing a small cost in the objective function for additional actions or resource consumption.} the chance that other agents (eg, humans, corporations, governments, or other AI systems) will use their agency to work against the first agent's objective is high, and it may therefore be rational to take steps or adopt a sub-goal to remove those actors' agency.

We believe that the emergence of instrumental subgoals is deeply connected to moral certainty. Agents that are not completely sure of the right thing to do (which we believe is an accurate summary of the state of knowledge about ethics, both because of normative impossibility and uncertainty theorems, and the practical difficulty of predicting the consequences of actions) are much more likely to tolerate the agency of others, than agents that are completely sure that they know the best way for events to unfold. This appears to be true not only of AI systems, but of human ideologies and politics, where totalitarianism has often been built on a substructure of purported moral certainty \cite{debeauvoir1947ambiguity,young1991totalitarian,hindy2015terrible}.

This leads us to propose a conjecture about the relationship between moral certainty and instrumentally convergent subgoals:

{\bf Totalitarian convergence conjecture:} {\em powerful agents with mathematically certain, monotonically increasing, open-ended objective functions will adopt sub-goals to disable or dis-empower other agents in all or almost all cases}.\footnote{The qualifier ``almost all'' requires some further specification. It requires that the objective not be finely tuned in some very intricate way that builds in preservation of all other agents. It is unclear if such fine-tuning is either properly definable or possible.}

A second conjecture is the converse of the first:

{\bf Pluralistic non-convergence conjecture:} {\em powerful agents with mathematically uncertain objectives will not adopt sub-goals to disable or dis-empower other agents unless those agents constitute a probable threat to a wide range of objectives.}




\subsection{Conclusion}

We have shown that impossibility theorems in ethics have implications for the design of powerful algorithmic systems, such as high-stakes AI applications or sufficiently rigid and rule-based bureaucracies. We showed that such paradoxes can be avoided by using uncertain objectives, such as partial orders or probability distributions over total orders; we proved uncertainty theorems that place a minimum bound on the amount of uncertainty required. Some previously proposed ethical theories (such Broome's vague neutral-level theory \cite[pp. 213-214]{broome2004weighing}) appear to satsify these bounds.

In the light of these results, we believe that machine learning researchers should avoid using totally ordered objective functions or loss functions as optimization goals in high-stakes applications. Systems designed that way appear to be suffering from an ethical ``type error'' in their goal selection (and action selection) code.

Instead, high-stakes systems should always exhibit uncertainty about the best action in some cases. Further study is warranted about the advantages of various probabilistic decision rules to handle such uncertainty, and about whether other mathematical models of uncertainy are better alternatives to the two models examined in this paper. Further research could also be productive on the relationship between ethical certainty and various observed and predicted pathological behaviour of AI systems. We proposed two conjectures on this topic, the Totalitarian Convergence Conjecture and the Pluralistic Non-Convergence Conjecture.

\section*{Acknowledgments}

The author would like to thank the late Aaron Swartz for having originally inspired this work and first attempted to make progress on this question; Toby Ord for many valuable conversations and a detailed framing of the problem; Rohin Shah for suggesting a simple proof strategy for the uncertainly satisfied bound; and Amanda Askell for providing valuable research assistance, and presenting an early poster version of this work at NIPS 2017, Daniel Ford for spotting issues and making valuable suggestions on the uncertain ordering bound. Thanks also to Dylan Hadfield-Menell, Laurent Orseau, Anca Dragan, and Stuart Armstrong for helpful suggestions and corrections on earlier versions of this work. This work was partially supported by a grant from the Open Philanthropy Project to EFF.  \newpage{\small \bibliography{papers}} \end{document}